\documentclass[letterpaper, 10 pt, journal, twoside]{IEEEtran}

\IEEEoverridecommandlockouts                              

\usepackage{graphicx} 
\usepackage{epsfig} 
\usepackage{amsmath} 
\usepackage{amssymb}  

\usepackage{float}

\usepackage{soul,color}
\usepackage[dvipsnames]{xcolor}

\usepackage{textcomp}

\usepackage{algorithm}
\usepackage{algorithmic}

\usepackage[caption=false,font=footnotesize,labelfont=sf,textfont=sf]{subfig} 
\usepackage{esvect} 
\usepackage{mathtools}

\newcommand{\mbf}[1]{\mathbf{#1}}

\newcommand{\mrf}[1]{\mathrm{#1}}

\newcommand{\pRb}{{^p\mbf{R}_b}}
\newcommand{\bRp}{{^b\mbf{R}_p}}

\newcommand{\bTp}{{^b\mbf{T}_p}}
\newcommand{\btp}{{^b\mbf{t}_p}}
\newcommand{\ptb}{{^p\mbf{t}_b}}
\newcommand{\pRc}{{^p\mbf{R}_c}}
\newcommand{\pTc}{{^p\mbf{T}_c}}
\newcommand{\ptc}{{^p\mbf{t}_c}}

\newcommand{\cTo}{{^c\mbf{T}_o}}

\newcommand{\F}[1]{\mathcal{F}_{#1}}
\newcommand{\dl}{\dot{\mbf{l}}}

\newcommand{\pui}{{^p\mbf{u}_i}}
\newcommand{\puone}{{^p\mbf{u}_1}}
\newcommand{\pum}{{^p\mbf{u}_m}}
\newcommand{\puonet}{{^p\mbf{u}_1^T}}
\newcommand{\pumt}{{^p\mbf{u}_m^T}}

\newcommand{\pAiBi}{{^p\vv{A_iB_i}}}

\newcommand{\AiBi}{A_iB_i}

\newcommand{\pai}{{^p\mbf{a}_i}}
\newcommand{\bai}{{^b\mbf{a}_i}}
\newcommand{\pbi}{{^p\mbf{b}_i}}
\newcommand{\pbone}{{^p\mbf{b}_1}}
\newcommand{\pbm}{{^p\mbf{b}_m}}

\newcommand{\de}{\dot{\mbf{e}}}
\newcommand{\I}[1]{\mbf{I}_{#1}}
\newcommand{\sskew}[1]{\lbrack {#1} \rbrack_\times }
\newcommand{\Ls}{\mbf{L_s}}
\newcommand{\Lshat}{\widehat{\mbf{L}}_{\mbf{s}}}
\newcommand{\Lw}{\mbf{L}_\omega}

\newcommand{\cvc}{{^c\mbf{v}_c}}
\newcommand{\sinc}[1]{\mrf{sinc}({#1})}
\newcommand{\sinctwo}[1]{\mrf{sinc}^2({#1})}
\newcommand{\hatw}[1]{\widehat{#1}}

\newcommand{\pvp}{{^p\mbf{v}_p}}

\newcommand{\Ad}{\mbf{A}_d}
\newcommand{\Adhat}{\widehat{\mbf{A}}_d}
\newcommand{\Ahat}{\widehat{\mbf{A}}}
\newcommand{\zeros}[1]{\mbf{0}_{#1}}

\renewcommand{\sin}[1]{\mrf{sin}({#1})}

\DeclarePairedDelimiterX{\norm}[1]{\lVert}{\rVert}{#1}

\newcommand{\bTphat}{{^b\mbf{\hatw{T}}_p}}
\newcommand{\pTchat}{{^p\mbf{\hatw{T}}_c}}

\newcommand{\twohalfdvs}{2\textonehalf D\,VS}

\title{
Robust 2 1/2D Visual Servoing of a Cable-Driven Parallel Robot Thanks to Trajectory Tracking
}

\author{Zane~Zake$^{1, 2}$,~\IEEEmembership{Student Member,~IEEE}, Fran\c{c}ois~Chaumette$^{3}$,~\IEEEmembership{Fellow,~IEEE},  Nicol\`{o}~Pedemonte$^{2}$, and St\'{e}phane~Caro$^{1,4}$,~\IEEEmembership{Member,~IEEE}
\thanks{Manuscript received: September, 10, 2019; Revised December, 5, 2019; Accepted January, 2, 2020.}
\thanks{This paper was recommended for publication by Editor Dezhen Song upon evaluation of the Associate Editor and Reviewers' comments.} 
\thanks{This work is supported by IRT Jules Verne (French Institute in Research and Technology in Advanced Manufacturing Technologies for Composite, Metallic and Hybrid Structures) in the framework of the PERFORM project.}
\thanks{$^{1}$Laboratoire des Sciences du Num\'{e}rique de Nantes, UMR CNRS 6004, 1, rue de la No\"{e}, 44321 Nantes, France,
{\tt\small Zane.Zake@ls2n.fr}}%
\thanks{$^{2}$IRT Jules Verne, Chemin du Chaffault, 44340, Bouguenais, France,
        {\tt\small nicolo.pedemonte@irt-jules-verne.fr}}%
\thanks{$^{3}$Inria, Univ Rennes, CNRS, IRISA, Rennes, France,
        {\tt\small Francois.Chaumette@inria.fr}}%
\thanks{$^{4}$Centre National de la Recherche Scientifique (CNRS), 1, rue de la No\"{e}, 44321 Nantes, France,
        {\tt\small stephane.caro@ls2n.fr}}%
\thanks{Digital Object Identifier (DOI): see top of this page.}
}

\begin{document}

\maketitle
 \pagestyle{empty}
 \thispagestyle{empty}

\markboth{IEEE Robotics and Automation Letters. Preprint Version. Accepted January, 2020}
{Zake \MakeLowercase{\textit{et al.}}: Robust 2\textonehalf D VISUAL SERVOING of A Cable-Driven Parallel Robot Thanks To Trajectory Tracking} 

\begin{abstract}

Cable-Driven Parallel Robots~(CDPRs) are a kind of parallel robots that have cables instead of rigid links. Implementing vision-based control on CDPRs leads to a good final accuracy despite modeling errors and other perturbations in the system. However, unlike final accuracy, the trajectory to the goal can be affected by the perturbations in the system. This paper proposes the use of trajectory tracking to improve the robustness of  {2\textonehalf D}~visual servoing control of CDPRs. 
Lyapunov stability analysis is performed and, as a result, a novel workspace, named control stability workspace, is defined. This workspace defines the set of moving-platform poses where the robot is able to execute its task while being stable. 
The improvement of robustness is clearly shown in experimental validation.
\end{abstract}



\section{INTRODUCTION}
\IEEEPARstart{A}{}
 special kind of parallel robots named Cable-Driven Parallel Robots~(CDPRs) has cables instead of rigid links. The main advantages of CDPRs are their large workspace, low mass in motion, high velocity and acceleration capacity, and reconfigurability~\cite{reconf}. The main drawback of CDPRs is their poor positioning accuracy. Multiple approaches to deal with this drawback can be found in the literature. The most common one is the improvement of the CDPR model. Since cables are not rigid bodies, creating a precise CDPR model is a tedious task, because it needs to include, for example, pulley kinematics, cable sag, elongation and creep~\cite{schmidt}~\cite{jpmerletICRA}~\cite{riehl}. Besides, cable-cable and cable-platform interferences can affect the accuracy of a CDPR. To avoid those interferences, studies have been done on the definition of CDPR workspace~\cite{reconf}. When modeling has been deemed unsuitable or insufficient, sensors have been used to gain knowledge about some of the system parameters. For example, angular sensors can be used to retrieve the cable angle~\cite{angular}; cable tension sensors can be used to assess the current payload mass and the location of center of gravity~\cite{picard}; color sensors can be used to detect regularly spaced color marks on cables to improve cable length measurement~\cite{jpmerletCableCon}. Of course, exteroceptive sensors can be used to measure the moving-platform~(MP) pose accurately. To the best of our knowledge, few studies exist on the use of vision to control CDPRs and improve their accuracy. For instance, four cameras are used in~\cite{dallej2019} to precisely detect the MP pose of a large-scale CDPR. Furthermore, additional stereo-camera pairs were used to detect cable sagging at their exit points from the CDPR base structure. Similarly,~\cite{chellal} used a six infra-red camera system to detect the MP pose of a CDPR used in a haptic application. A camera can also be mounted on the MP to see the object of interest. In this case, control is performed with respect to the object of interest. Thus the MP pose is not directly observed. Such a control algorithm for a three-DOF translational CDPR has been introduced in~\cite{remy}, and it has been extended to {six-DOF} CDPRs in~\cite{zakeICRA}, where the authors used a pose-based visual servoing~(PBVS) control scheme. The robustness of this control scheme to perturbations and uncertainties in the robot model was analyzed. The stability analysis of this controller was extended in~\cite{zakeCableCon} to find the limits of perturbations that do not yield the system unstable. As a conclusion, as long as the perturbations are kept within these limits, they do not affect the MP accuracy at its final pose. 
However, even if perturbation levels are kept within the boundaries, they have an undesirable effect along the trajectory to the~goal.

To further improve the robustness and the achievement of the expected trajectory, planning and tracking of a trajectory can be used. Trajectory planning and tracking take advantage of stability and robustness to large perturbations of classical visual servoing approaches in the vicinity of the goal~\cite{MezouarTRACK}. Indeed, when the difference between current and desired visual features is small, the behavior of the system approaches the ideal one, no matter the perturbations. With the implementation of trajectory planning and tracking, the desired features are varying along the planned trajectory keeping the difference between current and desired visual features small at all times.

Under perfect conditions, the PBVS control used in~\cite{zakeICRA} and~\cite{zakeCableCon} leads to a straight-line trajectory of the target center-point in the image, which means that the target is likely not to be lost during task execution. Unfortunately, even under perfect conditions the camera trajectory is not a straight line. To have a straight-line trajectory for both the target center-point in the image and the camera in the robot frame, a hybrid visual servoing control, named 2\textonehalf D~visual servoing~(\twohalfdvs)~\cite{2halfd}~\cite{2halfd_other}, has been selected in this paper. It combines the use of 2D and 3D features in order to profit from the benefits of PBVS and Image-Based Visual Servoing~(IBVS), while suffering the drawbacks of neither.

Accordingly, this paper deals with robust \twohalfdvs\ of a CDPR thanks to trajectory tracking. It allows us to ensure the predictability of the trajectory to the goal. Furthermore, it improves the overall robustness of the system. 
In addition, it was found in~\cite{zakeCableCon} that the stability of the system with given perturbations depends also on the MP pose in the base frame. As a consequence, a novel workspace, named Control Stability Workspace~(CSW), is defined. This workspace gives the set of MP poses where the robot is able to execute its task, while being stable from a control viewpoint.

This paper is organized as follows. 
Section~\ref{s2} presents the vision-based control strategy for a CDPR. Section~\ref{s3} is dedicated to the addition of trajectory planning and tracking in the control strategy. Stability of both control types is analyzed in Section~\ref{s4}. Section~V is dedicated to the definition of a novel workspace named Control Stability Workspace. Section~\ref{s5} describes the experimental results obtained on a small-scale CDPR. Finally, conclusions are drawn in Section~\ref{s6}.

\section{2\textonehalf D Visual Servoing of Cable-Driven Parallel Robots}
\label{s2}

\subsection{CDPR Kinematics}
\label{sub1}
\begin{figure}[thpb]
      \centering
      \includegraphics[width = 0.79\columnwidth]{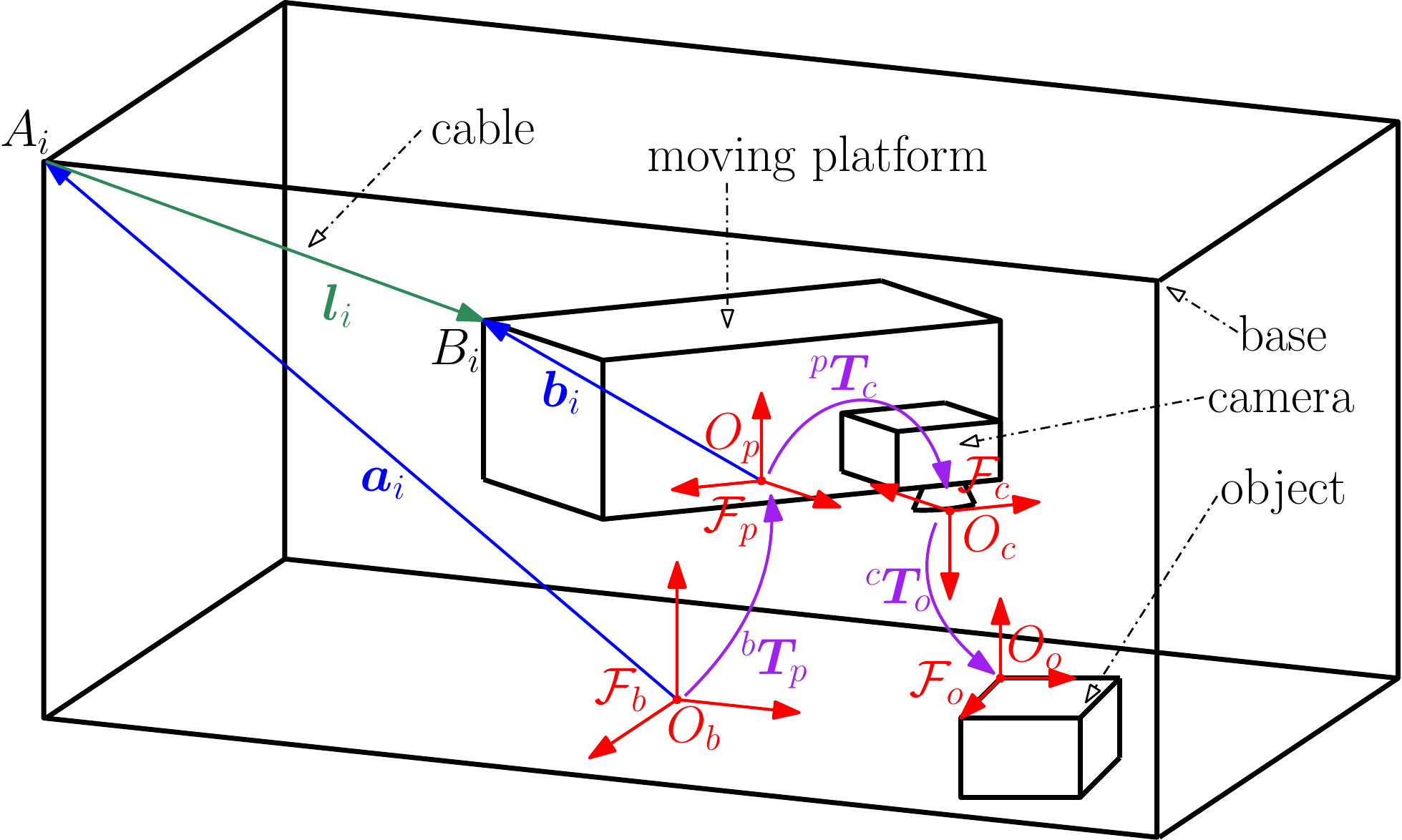}
      \caption{Schematic of a spatial CDPR with eight cables, a camera mounted on its MP and an object in the workspace}
      \label{fig:spatial_scheme}
   \end{figure}

The schematic of a spatial CDPR is shown in Fig.~\ref{fig:spatial_scheme}. The camera is mounted on the MP, therefore the homogeneous transformation matrix~$\pTc$ between the MP frame~$\F{p}$ and the camera frame~$\F{c}$ does not change with time. On the contrary, the homogeneous transformation matrices~$\bTp$ between the base frame~$\F{b}$ and the MP frame~$\F{p}$, and~$\cTo$ between the camera frame~$\F{c}$ and the object frame~$\F{o}$ change with time. 

The length $l_i$ of the $i$th cable is the 2-norm of the vector $\vv{\AiBi}$ pointing from cable exit point~$A_i$ to cable anchor point~$B_i$, namely,
\begin{equation} 
l_i =\norm[\big]{\vv{\AiBi}}_2
\end{equation}
with
\begin{equation} 
l_i \pui = \pAiBi = \pbi - \pai = \pbi - \pRb \bai - \ptb
\label{eq:l_long}
\end{equation}
where $\pui$ is the unit vector of $\pAiBi$ that is expressed as:
\begin{equation} 
\pui = \frac{\pAiBi}{\norm[\Big]{\pAiBi}_2}
=\frac{\pbi - \pai}{\norm[\Big]{\pAiBi}_2}
=\frac{ \pbi - \pRb \bai - \ptb}{\norm[\Big]{\pAiBi}_2}
\end{equation}
$\bai$ is the Cartesian coordinates vector of cable exit point~$A_i$ expressed in~$\F{b}$; $\pbi$ is the Cartesian coordinates vector of cable anchor point~$B_i$ expressed in~$\F{p}$; $\pRb$ and~$\ptb$ are the rotation matrix and translation vector from~$\F{p}$ to~$\F{b}$.

The cable velocities~$\dot{l}_i$ are obtained upon differentiation of Eq.~(\ref{eq:l_long}) with respect to~(w.r.t.) time:
\begin{equation} 
\dl = \mbf{A}  \pvp
\label{eq:kin}
\end{equation}
where~$\pvp$ is the MP twist expressed in its own frame~$\F{p}$, $\dl$~is the cable velocity vector, and~$\mbf{A}$ is the Forward Jacobian matrix of the CDPR,  defined~as~\cite{pott_book}:
\begin{equation} 
\mbf{A} =  \begin{bmatrix}
     \puonet & ( \pbone \times \puone)^T \\
     \vdots & \vdots \\
     \pumt & ( \pbm \times \pum)^T \\ 
     \end{bmatrix}
     \label{eq:structure}
\end{equation}
where~$m=8$ for a spatial CDPR with eight cables. Thus the Jacobian~$\mbf{A}$ is a $(8\times6)$--matrix.

\subsection{2\textonehalf D Visual Servoing}
\label{sub2}

\begin{figure}[thpb]
      \centering
      \includegraphics[width = 0.85\columnwidth]{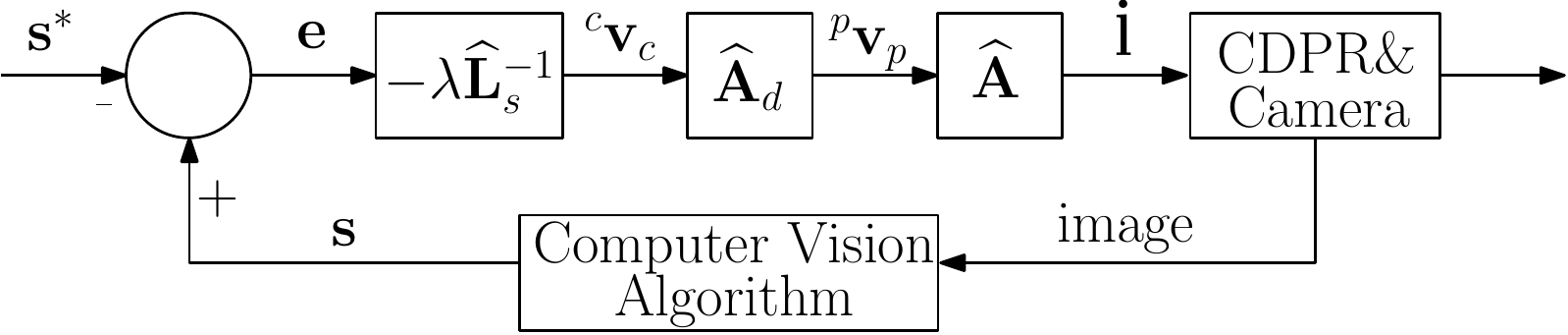}
      \caption{Control scheme for visual servoing of a CDPR}
      \label{fig:control}
   \end{figure}

The control scheme considered in this paper is shown in Fig.~\ref{fig:control}. An image is retrieved from the camera and processed with a computer vision algorithm, from which the current feature vector is defined as~$\mbf{s}=[{^{c^*}\!\mbf{t}_c^T} \ \ x_o \ \ y_o \ \ \theta u_z]^T$~\cite{2halfd}~\cite{2halfd_other}. Here,~${^{c^*}\!\mbf{t}_c}$ is the translation vector between the desired camera frame~$\F{c^*}$\footnote{In this paper, the superscript~$*$ denotes the desired value, e.g. desired feature vector~$\mbf{s}^*$. Similarly,~$c^*$ in~$^{c^*}\!\mbf{t}_c$ refers to desired camera frame~$\F{c^*}$} and the current camera frame~$\F{c}$;~$x_o$ and~$y_o$ are the image coordinates of the object center~$\mbf{o}$; $\theta u_z$ is the third component of~$\theta\mbf{u}$ vector, where~$\mbf{u}$ is the axis and~$\theta$ is the angle of the rotation matrix~${^{c^*}\!\mbf{R}_{c}}$. 
An error vector~$\mbf{e}$ is defined by comparing~$\mbf{s}$ to~$\mbf{s}^*$, namely 
\begin{equation}
\mbf{e} = \mbf{s}-\mbf{s}^*
\end{equation}

As mentioned in the introduction, in perfect conditions, this choice of visual features leads to a straight-line trajectory of the camera (because~${^{c^*}\!\mbf{t}_c}$ is part of~$\mbf{s}$), as well as a straight-line trajectory of object center-point~$\mbf{o}$ in the image (as~$(x_o,y_o)$ is also part of~$\mbf{s}$). The translational degrees of freedom are used to realize the 3D straight line of the camera, while the rotational degrees of freedom are devoted to the realization of the 2D straight line of point~$\mbf{o}$.

To decrease the error~$\mbf{e}$, an exponential decoupled form is selected 
\begin{equation}
\de = -\lambda \mbf{e}
\label{eq:de}
\end{equation}
with a positive adaptive gain~$\lambda$, that is computed at each iteration, depending on the current value of~$\norm{\mbf{e}}_2$~\cite{remy}. 
The derivative of the error~$\de$ can be written as a function of the Cartesian velocity of the camera~$\cvc$, expressed in~$\F{c}$:
\begin{equation}
\de = \Ls  \cvc
\label{eq:ecvc}
\end{equation}
where~$\Ls$ is the interaction matrix given by~\cite{2halfd}~\cite{2halfd_other}~\cite{chaumette_tut}:
\begin{equation}
\Ls= \begin{bmatrix}
   {^{c^*}\!\mbf{R}_{c}} & \zeros{3} \\
   \frac{1}{Z} \mbf{L}_{v} & \mbf{L}_{v\omega} \\
    \end{bmatrix}
    \label{eq_ls}
\end{equation}
with:
\begin{equation}
\mbf{L}_{v} = \begin{bmatrix} -1 & 0 & x_o \\ 0 & -1 & y_o \\ 0 & 0 & 0 \\
\end{bmatrix}
\end{equation}
\begin{equation}
\mbf{L}_{v\omega} = \begin{bmatrix}
x_oy_o & -(1+x_o^2) & y_o \\ (1+y_o^2) & -x_oy_o & -x_o \\ l_1 & l_2 & l_3 \\
\end{bmatrix}
\end{equation}
$l_1,l_2,l_3$ being the components of the third row of matrix~$\Lw$:
\begin{equation}
\Lw = \I{3} - \frac{\theta}{2}\sskew{\mbf{u}} + \Big(1- \frac{\sinc{\theta}}{\sinctwo{\frac{\theta}{2}}}\Big)\sskew{\mbf{u}}^2
\end{equation}
where $\sinc{\theta} = \sin{\theta}/\theta$

Finally, injecting~(\ref{eq:de}) into~(\ref{eq:ecvc}) the instantaneous velocity of the camera in its own frame can be expressed as:
\begin{equation}
\cvc = -\lambda \, \Lshat^{-1}\,\mbf{e}
\label{eq:cvc}
\end{equation}
where~$\Lshat^{-1}$ is the inverse of the estimation of the interaction matrix~$\Lshat$. Note that the inverse is directly used, because~$\Ls$ is a~$(6\times6)$--matrix that is of full rank for~\twohalfdvs~\cite{chaumette_tut}.

\subsection{Kinematics and Vision}

To control the CDPR by \twohalfdvs, it is necessary to combine the modeling shown in Sections~\ref{sub1} and~\ref{sub2}. It is done by expressing the MP twist~$\pvp$ as a function of camera velocity~$\cvc$:
\begin{equation}
\pvp = \Ad \cvc
\label{eq:bvp}
\end{equation}
where~$\Ad$ is the adjoint matrix that is expressed as~\cite{khalil}:
\begin{equation}
\Ad  =  \begin{bmatrix}
\pRc & \sskew{\ptc}\pRc\\ 
 \zeros{3} & \pRc\\
\end{bmatrix} 
\end{equation}

Finally, the model of the system shown in Fig.~\ref{fig:control} is written from Eqs.~(\ref{eq:kin}), (\ref{eq:ecvc}) and~(\ref{eq:bvp}): 
\begin{equation}
\de = \Ls \, \Ad^{\!-1} \mbf{A\!}^\dagger \, \dl
\label{eq:model}
\end{equation}
where~$\mbf{A\!}^\dagger$ is the Moore-Penrose pseudo-inverse of the Jacobian matrix~$\mbf{A}$.

Upon injecting~(\ref{eq:bvp}) and~(\ref{eq:cvc}) into~(\ref{eq:kin}), the output of the control scheme, i.e. the cable velocity vector~$\dl$, takes the form:
\begin{equation}
\dl =-\lambda \Ahat  \Adhat \Lshat^{-1}\mbf{e}
\label{eq:control}
\end{equation}
where~$\Ahat$ and~$\Adhat$ are the estimations of~$\mbf{A}$ and~$\Ad$, resp.

\section{Trajectory Planning and Tracking}
\label{s3}

It is well known that having perturbations in the system, which do not cause loss of stability, has an undesirable effect on the trajectory. This was shown in~\cite{zakeICRA}~\cite{zakeCableCon} for PBVS and it is also true for the \twohalfdvs\ controller.
Trajectory planning and following can be used to increase the robustness of the chosen control w.r.t. modeling errors~\cite{MezouarTRACK} and to preserve the straight-line shape of the trajectory~\cite{martinetTRACK}.

Indeed, the larger~$\mbf{e}=\mbf{s}-\mbf{s}^*$, the bigger the effect of modeling errors on system behavior. When tracking a chosen trajectory, at each iteration $i$ the error becomes ${\mbf{e}(t)=\mbf{s}(t)-\mbf{s}^*(t)}$. Consequently, when~$t=0$\,s we have ${\mbf{s}^*(0)=\mbf{s}(0)}$. Since~$\mbf{s}^*(t)$ is now time varying, the control scheme needs to be slightly changed. More precisely, instead of~(\ref{eq:ecvc}) we now have~\cite{chaumette_tut}~\cite{martinetTRACK}:
\begin{equation}
\de = \dot{\mbf{s}}- \dot{\mbf{s}}^* = \Ls \cvc -  \dot{\mbf{s}}^*
\label{eq:ecvc_track}
\end{equation}

\begin{figure}[thpb]
      \centering
      \includegraphics[width = 1\columnwidth]{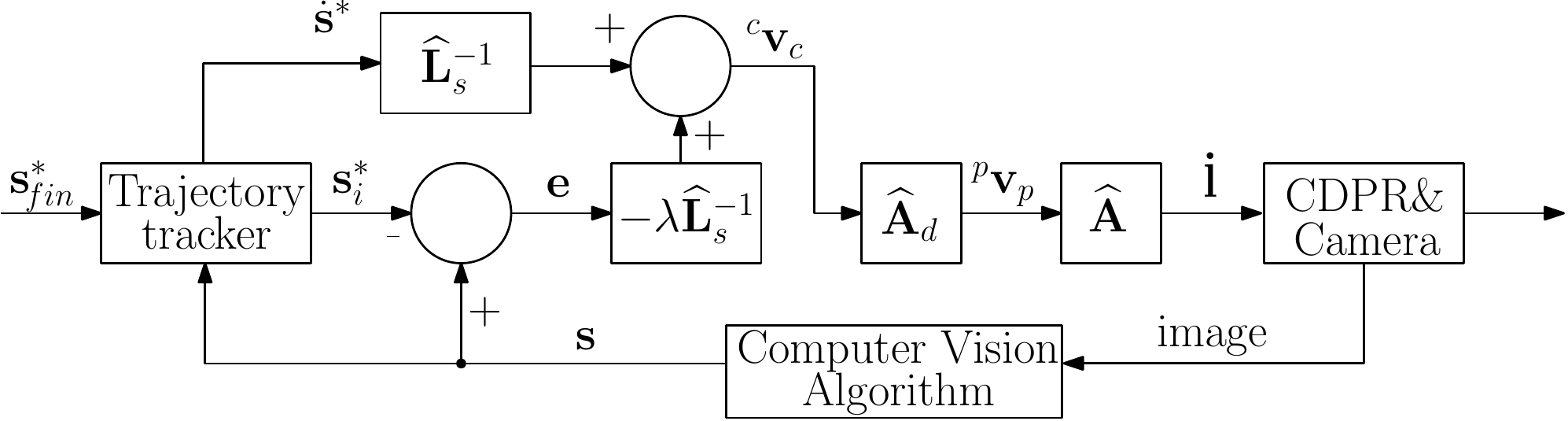}
      \caption{Control scheme for VS with trajectory tracking of a CDPR}
      \label{fig:controlTrajTrack}
   \end{figure}

Hence, the new control scheme is shown in Fig.~\ref{fig:controlTrajTrack}.

The new model of the system shown in Fig.~\ref{fig:controlTrajTrack} is written from Eqs.~(\ref{eq:kin}), (\ref{eq:ecvc_track}) and~(\ref{eq:bvp}):
\begin{equation}
\de = \Ls \Ad^{\!-1} \mbf{A\!}^\dagger \, \dl - \dot{\mbf{s}}^*
\label{eq:modelTrack}
\end{equation}

Injecting~(\ref{eq:bvp}) and~(\ref{eq:ecvc_track}) into~(\ref{eq:kin}) and expressing cable velocity vector~$\dl$ leads to :
\begin{equation}
\dl = \Ahat \Adhat \Lshat^{-1}(-\lambda \mbf{e} + \hatw{\dot{\mbf{s}}^*})
\label{eq:controlTrack}
\end{equation}
 
The success of any trajectory tracking is based on the time available to complete the task. The higher the trajectory time~$t_{full}$, the more accurate the trajectory tracking. Indeed, the larger~$t_{full}$, the lower the MP velocity, and the smaller the path step between two iterations. This leads to a smaller difference between~$\mbf{s}^*(t)$ and~$\mbf{s}^*(t+\Delta t)$, which in turn means a smaller difference between~$\mbf{s}(t)$ and~$\mbf{s}^*(t+\Delta t)$, thus a better path following.

\subsection{Implementation for~\twohalfdvs}

The implementation of the trajectory planning and  tracking for  \twohalfdvs\ is shown in Algorithm~1.
There are three distinct phases, the first being the initialization, the second being the trajectory planning, and the third being the trajectory tracking. During the initialization phase, the final desired object pose~$^{c^*}\mbf{\!p}^*_{o_{fin}}$ and center-point~$\mbf{o}^*_{fin}$ are defined. They are used to compute the final feature vector~$\mbf{s}^*_{fin}$. Similarly, the initial feature vector~$\mbf{s}_{init}$ is defined based on the initial pose~$^c\mbf{p}_{o_{init}}$ and center-point~$\mbf{o}_{init}$ of the object of interest that are measured and recorded. This allows us to compute the full error: 
\begin{equation}
{\mbf{e}_{full} = \mbf{s}_{init} - \mbf{s}^*_{fin}}
\label{eq:efull}
\end{equation}
and trajectory time:
\begin{equation}
t_{full} = \text{max}(\frac{\text{e}_n}{\text{v}_n}, n = 1 \ \text{to}\ 6)
\label{eq:time}
\end{equation}
where~$\text{e}_n$ stands for the~$n$-th component of~$\mbf{e}_{full}$; $\text{v}_n$ stands for the~$n$-th component of the desired average velocity~$\mbf{v}$.

The current desired feature vector~$\mbf{s}^*(t)$ varies at a constant velocity~$\mbf{c}$ that is expressed as:
\begin{equation}
\mbf{c} = \frac{\mbf{e}_{full}}{t_{full}}
\label{eq:c}
\end{equation}

At the trajectory planning phase, we define~$\mbf{s}^*(t)$. At the beginning, when~$t=0$\,s, it is clear that ${\mbf{s}^*(0)=\mbf{s}_{init}}$. Then for~$i=1,\hdots,k$, where~$k=t_{full}/\Delta t$ and~$\Delta t$ is the time interval between two iterations, 
the trajectory planning is expressed as:
\begin{equation}
\mbf{s}^*(i\Delta t)=\mbf{s}_{init}+i  \Delta t \,\mbf{c}
\end{equation}

As a consequence, we can set in~(\ref{eq:controlTrack}):
\begin{equation}
\begin{cases}
\dot{\mbf{s}}^* = \hatw{\dot{\mbf{s}}^*} = \mbf{c}\  \text{when}\  t<t_{full} \\
\dot{\mbf{s}}^* = \hatw{\dot{\mbf{s}}^*} = \mbf{0} \  \text{when}\  t\geq t_{full} \\
\end{cases}
\label{equalities}
\end{equation}

The third phase iterates until the difference~$\norm[\big]{\mbf{s}(t)-\mbf{s}^*_{fin}}_2$ reaches a defined threshold.  At each iteration, the current feature vector~$\mbf{s}(t)$ is computed from the current object pose and the current object center-point coordinates. The current desired feature vector~$\mbf{s}^*(t)$ is retrieved from trajectory planning algorithm.
This allows us to compute the current error~${\mbf{e}(t) = \mbf{s}(t)-\mbf{s}^*(t)}$, which is then used as input of the control scheme.

\begin{algorithm}[h]
\floatname{algorithm}{Algorithm 1:}
\renewcommand{\thealgorithm}{}
\caption{Trajectory planning and tracking}
\label{algo1}
\renewcommand\algorithmicrepeat{\textbf{Initialization}}
\renewcommand\algorithmicuntil{\textbf{End of Initialization}}
\begin{algorithmic}[1]
\newcommand{\INDSTATE}[1][1]{\STATE\hspace{#1\algorithmicindent}}

\REPEAT
\STATE Set the desired object pose $^{c^*}\!\mbf{p}^*_{o_{fin}}$ and center-point coordinates $\mbf{o}_{fin}^*$
\STATE Define final feature vector $\mbf{s}^*_{fin}$
\STATE Read and record initial object pose~$^c\mbf{p}_{o_{init}}$ and center-point coordinates~$\mbf{o}_{init}$
\STATE Define initial feature vector  $\mbf{s}_{init}$
\STATE Compute trajectory time $t_{full}$ from~(\ref{eq:time})
\STATE Compute the constant velocity $\mbf{c}$ as in (\ref{eq:c})
\UNTIL{}
\renewcommand\algorithmicrepeat{\textbf{Trajectory Planning}}
\renewcommand\algorithmicuntil{\textbf{End of Trajectory Planning}}
\renewcommand\algorithmicdo{\textbf{record}}
\STATE
\REPEAT
\STATE $\mbf{s}^*(0) = \mbf{s}_{init}$
\STATE $k=t_{full}/\Delta t$
\FOR{$i = 1:k$}
\STATE $\mbf{s}^*(i\Delta t) = \mbf{s}_{init}+i  \Delta t\, \mbf{c}$
\ENDFOR
\UNTIL{}
\renewcommand\algorithmicrepeat{\textbf{Trajectory Tracking}}
\renewcommand\algorithmicuntil{\textbf{End of Trajectory Tracking}}
\renewcommand\algorithmicdo{\textbf{do}}
\STATE
\REPEAT
\WHILE{$\norm[\big]{\mbf{s}(t)-\mbf{s}^*_{fin}}_2>\text{threshold}$}
	\STATE Retrieve current desired feature vector $\mbf{s}^*(t)$
	\STATE Compute current feature vector $\mbf{s}(t)$ 
	\STATE Compute current error $\mbf{e}(t)=\mbf{s}(t)-\mbf{s}^*(t)$
	\STATE Compute current $\Lshat$, $\Ahat$ and $\Adhat$
	\STATE Compute $\dl$ using~(\ref{eq:controlTrack}) and send to CDPR
\ENDWHILE
\UNTIL{}
\end{algorithmic}
\end{algorithm}

\section{Stability Analysis}
\label{s4}

The ability of a system to successfully complete its tasks can be characterized by its stability. By analyzing system stability, it is possible to find the limits of perturbation on different variables that the system is able to withstand, that is, to determine whether the system is able to converge accurately to its goal despite the perturbations~\cite{khalil_syst}. 

In this paper, Lyapunov analysis is used to determine the stability of the closed-loop system.

\subsection{2\textonehalf D Visual Servoing}

The following closed-loop equation is obtained from~(\ref{eq:model}) and~(\ref{eq:control}):
\begin{equation}
\de =-\lambda  \Ls \Ad^{\!-1} \mbf{A\!}^\dagger \Ahat \Adhat \Lshat^{-1}\mbf{e}
\label{eq:closed}
\end{equation}

From (\ref{eq:closed}), a sufficient condition to ensure the system stability is~\cite{khalil_syst}:
\begin{equation}
\mbf{\Pi} = \Ls \Ad^{\!-1} \mbf{A\!}^\dagger  \Ahat \Adhat \Lshat^{-1} > \mbf{0} , \forall t
\label{eq:crit}
\end{equation}

Indeed, if this condition is satisfied, the error~$\mbf{e}$ will always decrease to finally reach~$\mbf{0}$. 

\subsection{Trajectory tracking with 2\textonehalf D Visual Servoing}

When trajectory tracking is involved, the closed-loop equation is written by injecting~(\ref{eq:controlTrack}) into~(\ref{eq:modelTrack}). Then, by using~(\ref{equalities}), we obtain:
\begin{equation}
\de = -\lambda \Ls \Ad^{\!-1} \mbf{A\!}^\dagger \Ahat \Adhat \Lshat^{\!-1}\mbf{e} + \Ls \Ad^{\!-1} \mbf{A\!}^\dagger \Ahat \Adhat \Lshat^{\!-1}\dot{\mbf{s}}^* -  \dot{\mbf{s}}^*
\label{eq:closedTrack2}
\end{equation}

The stability criterion~$\mbf{\Pi}$ keeps the form defined in~(\ref{eq:crit}). However, even if~$\mbf{\Pi}$ is positive definite, the error~$\mbf{e}$ will decrease iff the estimations 
are sufficiently accurate so that
\begin{equation}
\Ls \Ad^{\!-1} \mbf{A\!}^\dagger \Ahat \Adhat \Lshat^{\!-1}\dot{\mbf{s}}^* \approx \dot{\mbf{s}}^*
\label{eq:trackErr}
\end{equation}

Otherwise tracking errors will be observed. This can be explained by a simple example from~\cite{chaumette_tut}, where a scalar differential equation~$\dot{e} = -\lambda e + b$, which is a simplification of~(\ref{eq:closedTrack2}), is analyzed. The solution is~${e(t)=e(0)\text{exp}(-\lambda t) + b/\lambda}$, which converges towards~$b/\lambda$. Increasing~$\lambda$ reduces the tracking error. However, if it is too high, it can yield the system unstable. Therefore, it is necessary to keep~$b$ as small as possible.

Most importantly, as the current desired feature vector~$\mbf{s}^*(t)$ approaches regularly the final desired feature vector~$\mbf{s}^*$, the desired feature vector velocity~$\dot{\mbf{s}}^*$ will become~$\mbf{0}$ as stated in~(\ref{equalities}), which makes the tracking errors vanish at the end.

\section{Control Stability Workspace}

Before using a CDPR, one needs to know its workspace. Among the existing workspaces~\cite{stump}~\cite{verhoeven}, the static feasible workspace~(SFW) is the simplest one and is formally expressed as~\cite{reconf}:
\begin{equation}
\mathcal{F} = \{ \mbf{p}_p \in {SE(3):} \ \exists\boldsymbol{\tau}\in\mathcal{T},\ \mbf{W}\boldsymbol{\tau}+\mbf{w}_{g}=\zeros{6}\}
\end{equation}
Namely, the workspace~$\mathcal{F}$ is the set of all MP poses~$\mbf{p}_p$ for which there exists a vector of cable tensions~$\boldsymbol{\tau}$ within the cable tension space~$\mathcal{T}$ such that the CDPR can balance the gravity wrench $\mbf{w}_g$, and~$\mbf{W}\boldsymbol{\tau}+\mbf{w}_g=\zeros{6}$. Here,~$\mbf{W}$ is the wrench matrix and it is related to the robot Jacobian as~$\mbf{W} = -\mbf{A\!}^T$.

This workspace is a kineto-static workspace that shows all the poses that the MP is physically able to attain. In addition, it is important to evaluate the CDPR ability to reach a pose from a control perspective.

In~\cite{zakeCableCon} it was concluded that the results of stability analysis were dependent on the size of the MP workspace. The smaller the desired workspace, the larger the tolerated perturbations within system stability. The MP pose and stability analysis are related to each other, because the MP pose shows up in the stability criterion~$\mbf{\Pi}$ through the Jacobian matrix~$\mbf{A}$ in the form of rotation matrix~$\bRp$ and translation vector~$\btp$.

According to the stability analysis of \twohalfdvs\ control, presented in Section~\ref{s4}, the corresponding workspace, named Control Stability Workspace~(CSW), is defined as follows:
\begin{equation}
\mathcal{C} = \{\mbf{p}_p \in  SE(3) : \forall \mbf{d} \in \mathcal{D}, \mbf{\Pi}> \mbf{0}\}
\end{equation}
The workspace~$\mathcal{C}$ is the set of all MP poses~$\mbf{p}_p$, for which the stability criterion~$\mbf{\Pi}$ is positive definite for any vector of perturbations~$\mbf{d}$ that is within bounds~$\mathcal{D}$.
It means that for any MP pose within its CSW, the robot controller will be able to guide the MP to its goal.

It is of interest to create a compound workspace, that takes into account the controller and the kineto-static performance of the robot. 
Indeed, on the one hand, a MP pose can belong to~$\mathcal{F}$ while being outside of~$\mathcal{C}$, namely, it is in a static equilibrium, but it will fail to reach the goal. On the other hand, a MP pose can belong to~$\mathcal{C}$ while being outside of~$\mathcal{F}$, namely, the robot controller will make the MP reach the goal although the MP is not in a static equilibrium.
Thus we define a compound workspace, named $\mathcal{FC}$, as the intersection of~$\mathcal{F}$ and~$\mathcal{C}$:
\begin{IEEEeqnarray}{cC}
\mathcal{FC}=\{ \mbf{p}_p \in SE(3): \exists\boldsymbol{\tau}\in\mathcal{T},\ \forall\mbf{d} \in \mathcal{D},  \IEEEnonumber \\   \ \mbf{W}\boldsymbol{\tau}+\mbf{w}_{g}=\zeros{6}, \ \mbf{\Pi}>\mbf{0}\}
\end{IEEEeqnarray}
The compound workspace~$\mathcal{FC}$ is the set of all MP poses~$\mbf{p}_p$ for which there exists a vector of cable tensions~$\boldsymbol{\tau}$ within the cable tension space~$\mathcal{T}$ such that the CDPR can balance the gravity wrench $\mbf{w}_g$ leading to~$\mbf{W}\boldsymbol{\tau}+\mbf{w}_g=\zeros{6}$, and for which for any vector of perturbations~$\mbf{d}$ that is within bounds~$\mathcal{D}$, the stability criterion~$\mbf{\Pi}$ is positive definite.

\section{Experimental Setup and Validation}
\label{s5}

Stability criterion~(\ref{eq:crit}) 
is robot model dependent. Thus, ACROBOT, the CDPR prototype used for experimental validation, is presented in~Section~\ref{acro}.
Workspace $\mathcal{C}$~is computed in Section~\ref{numerical-analysis} based on the numerical analysis of the stability criterion~(\ref{eq:crit}).
Finally, experimental results are shown in Section~\ref{experiments}.

\subsection{CDPR prototype ACROBOT}
\label{acro}

\begin{figure}[thpb]
    \centering
    \includegraphics[width = 0.99\columnwidth]{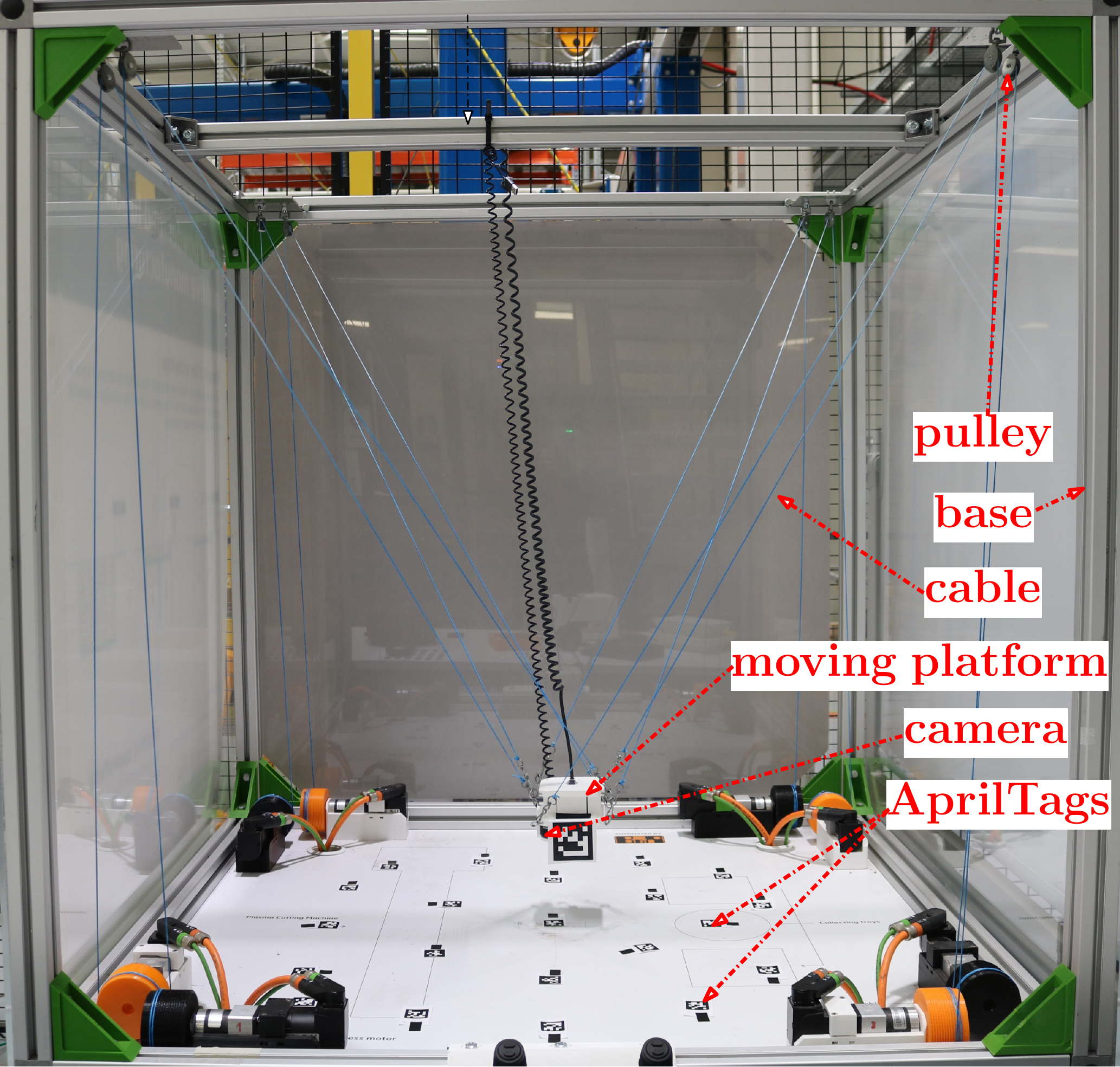}
    \caption{ACROBOT: a CDPR prototype located at IRT Jules Verne, Nantes}
    \label{fig:acrobot}
\end{figure}

CDPR prototype ACROBOT is shown in Fig.~\ref{fig:acrobot}. It is assembled in a suspended configuration, so that all the cable exit points are located at the four corners above the MP.  Cables are 1.5\,mm in diameter, assumed to be massless and nonelastic. The frame of the robot is a 1.2\,m\,$\times$\,1.2\,m\,$\times$\,1.2\,m cube. The MP size is 0.1\,m\,$\times$\,0.1\,m\,$\times$\,0.07\,m and its mass is~1.5\,kg.
   
A camera is mounted on the MP facing the ground. As a simplification of the vision part, AprilTags~\cite{apriltag} are used as objects and are put in various places on the ground. Their recognition and localization are done by algorithms available in the ViSP library~\cite{visp}. The robot is controlled to arrive directly above a chosen AprilTag.

\subsection{Constant and varying perturbations in the system}
\label{perturbations}
Two types of perturbations are considered depending on whether they change during task execution or not. Here is a list of perturbed parameters that do not change during the task execution:
	\begin{itemize}
	\item $\pTchat$ - the pose of the camera in the MP frame~$\F{p}$ can be perturbed due to hand-eye calibration errors. It affects the adjoint matrix $\Adhat$;
	\item $^p\hat{\mbf{b}}_i$ - the Cartesian coordinates vector of cable anchor points expressed in~$\F{p}$ can be perturbed due to manufacturing errors. It affects the estimation of Jacobian matrix~$\Ahat$;
	\item $^b\hat{\mbf{a}}_i$ - the Cartesian coordinates vector of cable exit points expressed in $\F{b}$. Since pulleys are not modeled, there is a small difference between the modeled and the actual cable exit points. It affects the estimation of Jacobian matrix~$\Ahat$.
	\end{itemize}

\begin{figure*}[!h]%
\centering
\subfloat[\label{wsa}]{\includegraphics[width = 0.33\textwidth]{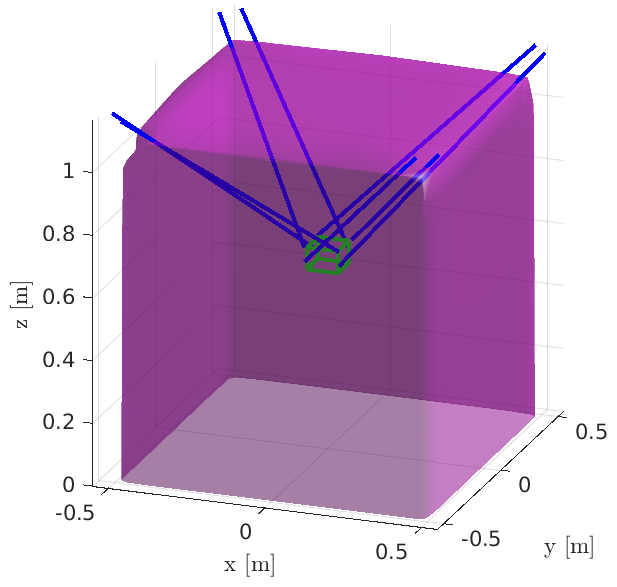}}%
\hfil 
\subfloat[\label{wsb}]{\includegraphics[width = 0.33\textwidth]{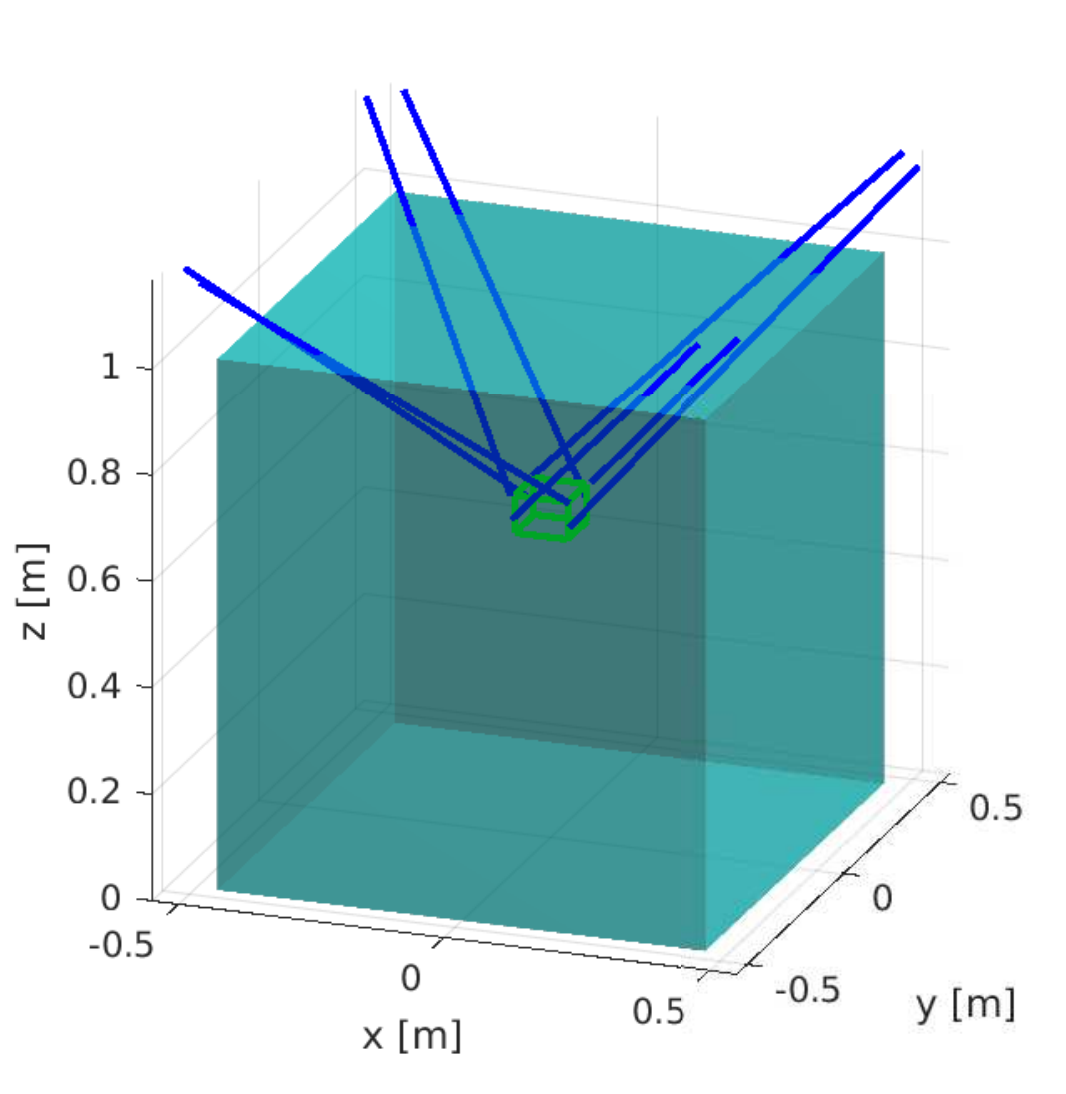}}%
\hfil
\subfloat[\label{wsc}]{\includegraphics[width = 0.33\textwidth]{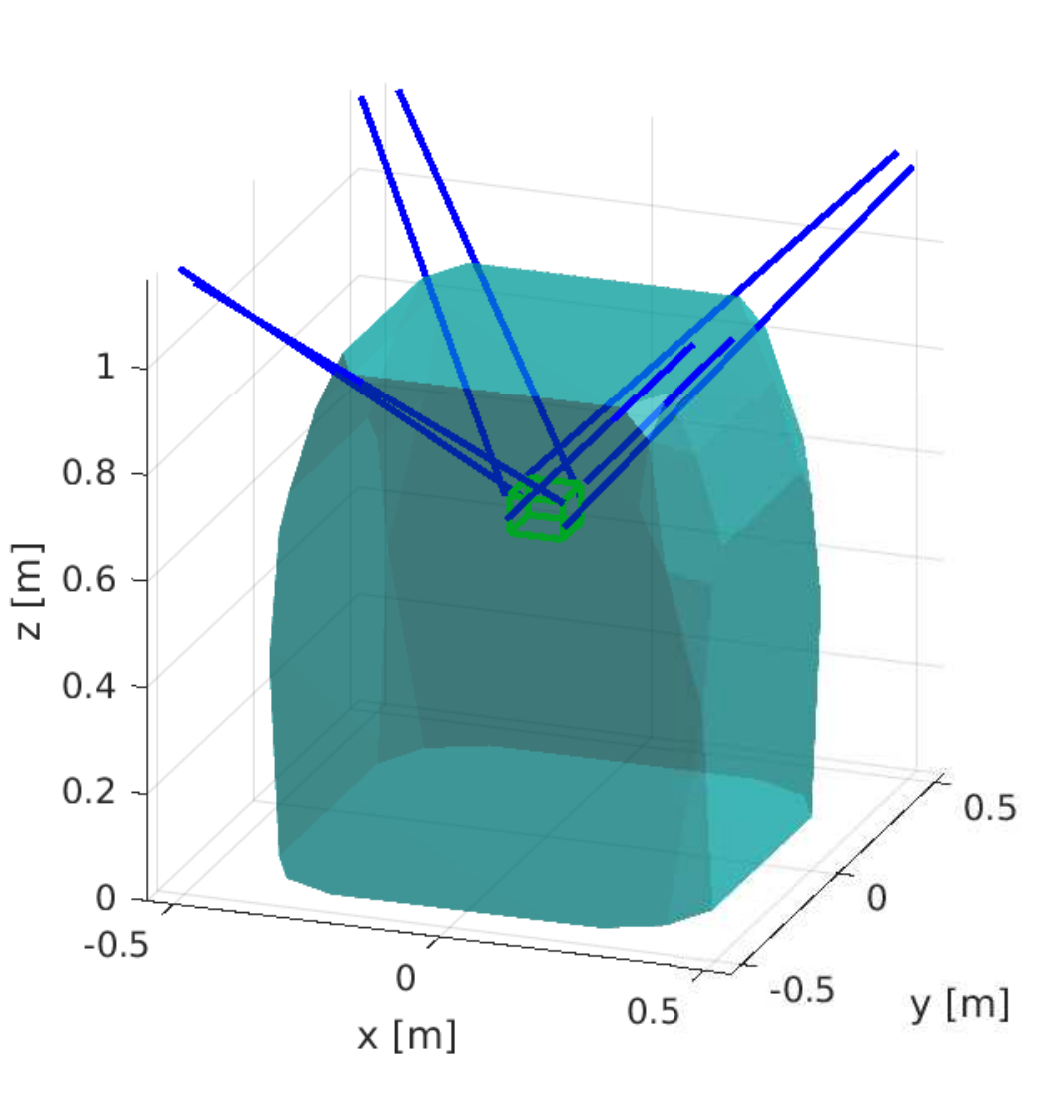}}%
 \caption{Workspace visualizations for ACROBOT: a)~SFW; b)~CSW for \twohalfdvs\ with minimal perturbations in the system and constant MP orientation; c)~CSW for \twohalfdvs\ with non-negligible perturbations in the system and MP rotation up to 30$^\circ$ about any arbitrary axis. }
\label{fig:figuresss2}
\end{figure*}	
	
Here is a list of the perturbed parameters that vary during the task execution:
	\begin{itemize}
	\item ${\mbf{s}}$ - the feature vector requires current AprilTag Cartesian pose in~$\F{c}$ and the image coordinates of its center-point~$\mbf{o}$. Those terms are computed from image features and are thus corrupted by noise. The smaller the AprilTag in the image, the larger the estimation error. It affects the interaction matrix~$\Lshat$;
	\item $\bTphat$ - the transformation matrix between frames~$\F{b}$ and~$\F{p}$ is estimated by exponential mapping:
	\begin{equation}
	(\bTp)_{i+1} = (\bTp)_i \,\text{exp}^{(\pvp, \Delta t)}
	\end{equation}
Since~$\pvp$ is computed from~$\cvc$, which is perturbed by errors in~$\pTchat$, and since computed~$\cvc$ does not correspond exactly to achieved~$\cvc$ due to errors in~$\Ahat$ and due to the time-response of the low-level controller, 
then~$\bTphat\neq\bTp$. Furthermore, the initial position is only coarsely known\footnote{The knowledge of initial MP pose is usually difficult to acquire when working with CDPRs. The usual approach is to always finish a task at a known home pose. This can be impossible due to a failed experiment or an emergency stop. Furthermore, great care must be taken when measuring the home pose, which in case of ACROBOT was done by hand.}. It affects the Jacobian matrix~$\Ahat$.

	\end{itemize}

\subsection{Workspace of ACROBOT}
\label{numerical-analysis}

The constant orientation static feasible workspace of ACROBOT was traced thanks to ARACHNIS software~\cite{arachnis} and is shown in Fig.~\ref{wsa}. 

CSW for ACROBOT is shown in Fig.~\ref{wsb}. Here, for the sake of comparison we also constrain the MP to the same constant orientation. 
Furthermore, we also take into account hand-eye calibration errors in camera pose in the MP frame~$\F{p}$, which are simulated as 0.01\,m along and 3$^\circ$ about any arbitrary axis. Finally, the MP pose is assumed to be estimated coarsely, allowing for an error of 0.05\,m in translation and 10$^\circ$ in rotation along and about any arbitrary axis. 

Figure~\ref{wsc} shows a smaller CSW, where the system will remain stable with non-negligible perturbations. Namely, we add 0.19\,m translational error and 8.5$^\circ$ rotational error along and about any arbitrary axis to the initial MP pose. Furthermore, we also simulate a bad hand-eye calibration by adding a 30$^\circ$ error to the camera pose in $\F{p}$. Finally, since we are interested in changing the orientation of the MP, CSW shown in Fig.~\ref{wsc} allows for up to~$\pm30^\circ$ rotation of the MP about any arbitrary axis.
   
\subsection{Experimental Validation}
\label{experiments}
An experimental setup was designed to validate the proposed approach. 
For \twohalfdvs\ we used an adaptive gain~$\lambda$~\cite{remy}:
\begin{equation}
\lambda(x) = (\lambda_0-\lambda_\infty)\textrm{e}^{-(\dot{\lambda}_0/(\lambda_0-\lambda_\infty))x}+\lambda_\infty
\end{equation}
where:
\begin{itemize}
\item $x = ||\mbf{e}||_2$ is the 2--norm of error $\mbf{e}$ at the current iteration 
\item $\lambda_0 = \lambda(0)$ is the gain tuned for very small values of $x$
\item $\lambda_\infty = \lambda(\infty)$ is the gain tuned for very high values of~$x$
\item $\dot{\lambda}_0$ is the slope of $\lambda$ at  $x=0$
\end{itemize}

These coefficients have been tuned at the following values: ${\lambda_0=2.0}$, ${\lambda_\infty=0.4}$ and ${\dot{\lambda}=30}$. 
 
For the controller with trajectory tracker, $\lambda=\lambda_0=2.0$ has been set, since the error is always small. Additionally, for the planner $t_{full}$ is set to be equal to the execution time of the classic \twohalfdvs\ in order to ease comparability of the results. Finally, $\Delta t=0.05$\,s.
 
The initial values are the following:
$$
\begin{cases}
^b\mbf{p}_{p} \!= \!\big[0.107\,m; \  -0.026\,m; \ 0.35\,m; \ +20^\circ; \ -20^\circ; \ 0^\circ\big] \\
^c\mbf{p}_{o}\!=\!\big[\!-\!0.022\,m;  0.136\,m;  0.449\,m;  -157^\circ;  -18^\circ;  -176^\circ\big]\\
 \mbf{o}=\big[-0.043\,m; \ 0.301\,m\big] \\
\end{cases}
$$
and final desired values are selected to be:
$$
\begin{cases}
^b\mbf{p}_{p}^* = \big[0.30\,m; \ 0.25\,m; \ 0.12\,m; \ 0^\circ; \ 0^\circ; \ 0^\circ\big] \\
^c\mbf{p}_{o}^*=\big[0\,m; \ 0\,m; \ 0.09\,m; \ -180^\circ; \ 0^\circ; \ -180^\circ\big] \\
 \mbf{o}^*=\big[0\,m; \ 0\,m\big] \\
\end{cases}
$$
where~$^b\mbf{p}_{p}$ denotes the MP pose in the base frame~$\F{b}$; $^c\mbf{p}_{o}$ denotes the AprilTag pose in the camera frame~$\F{c}$; and~$\mbf{o}$ stands for the AprilTag center-point coordinates in the image. Note that~$^c\mbf{p}_{o}$ and~$\mbf{o}$ were measured, while~$^b\mbf{p}_{p}$ was estimated through the $\bTp$ exponential mapping explained in Section~\ref{perturbations}. Therefore, the $^b\mbf{p}_{p}$ and $^b\mbf{p}_{p}^*$ are shown as a reference to Fig.~\ref{wsc}, but are not used in the~control.

Two perturbation sets are defined as $V1$ and $V2$. The former corresponds to the CSW shown in Fig.~\ref{wsc} and includes: $(i)$ a perturbation of initial MP pose of 0.19\,m along axis ${\mbf{u}=\big[0.56; 0.64; 0.52\big]}$ and $8.4^\circ$ about axis ${\mbf{u}=\big[0.73; 0.67; -0.14\big]}$; $(ii)$ and a perturbation on the camera orientation expressed in~$\F{p}$ of~$18^\circ$ about axis ${\mbf{u}=\big[0.61; -0.51; -0.61\big]}$. The set $V2$ includes: $(i)$ a perturbation of initial MP pose of 0.13\,m along axis ${\mbf{u}=\big[-\!0.57; -0.52; 0.63\big]}$ and 9.5$^\circ$ about axis ${\mbf{u}=\big[-\!0.52; 0.85; -0.04\big]}$; $(ii)$ a perturbation of camera pose in $\F{p}$ of 0.05\,m along y axis and 12.5$^\circ$ about axis ${\mbf{u}=\big[0.78; -0.51; -0.35\big]}$; $(iii)$ and a perturbation of 0.005\,m in a random direction for each cable exit point~$A_i$ and anchor point~$B_i$.

\begin{figure*}[!h]%
\centering
\subfloat[\label{suba}]{\includegraphics[height = 6cm]{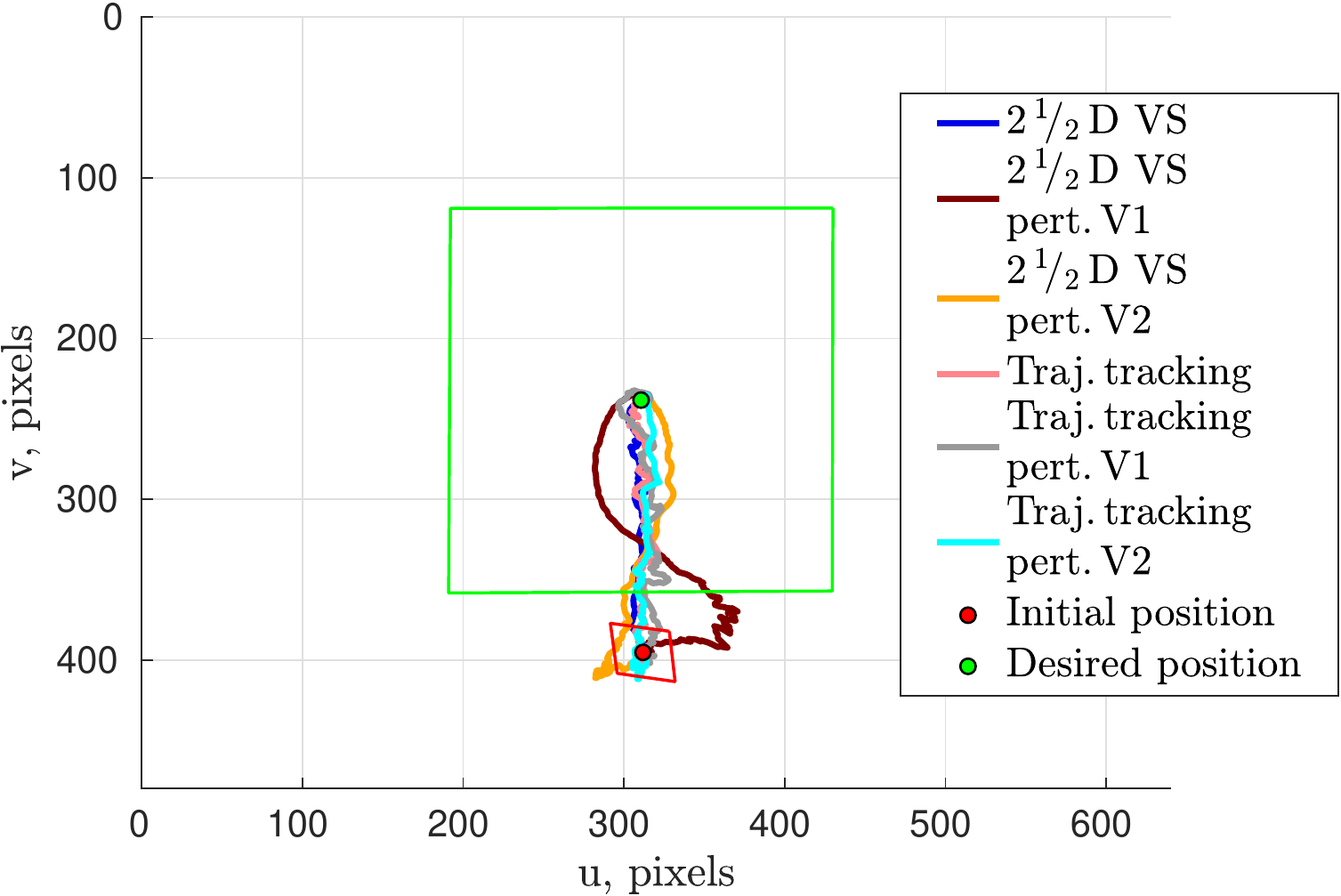}}%
\hfil 
\subfloat[\label{subb}]{\includegraphics[height = 6cm]{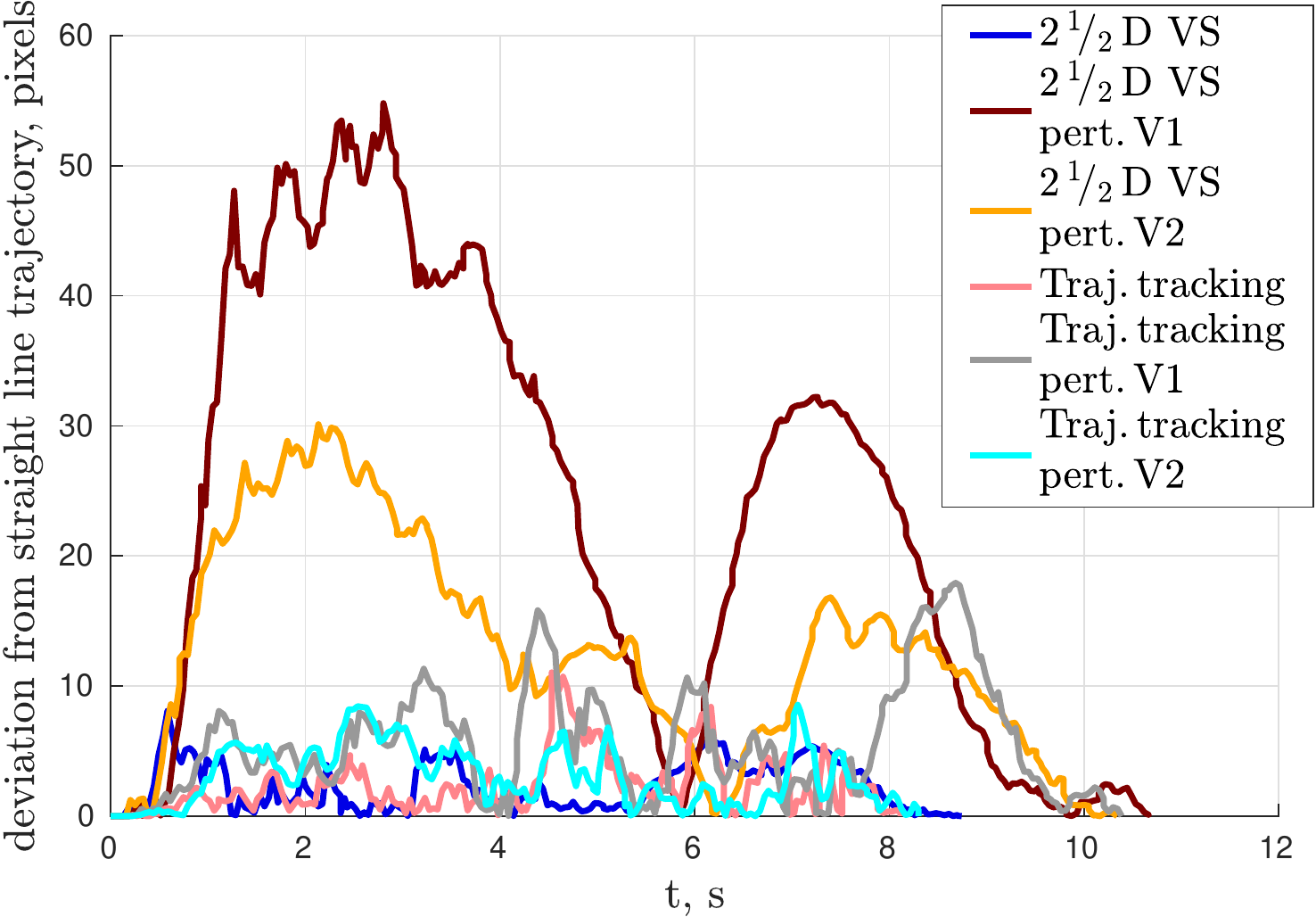}}%
\hfil
\subfloat[\label{subc}]{\includegraphics[height = 6cm]{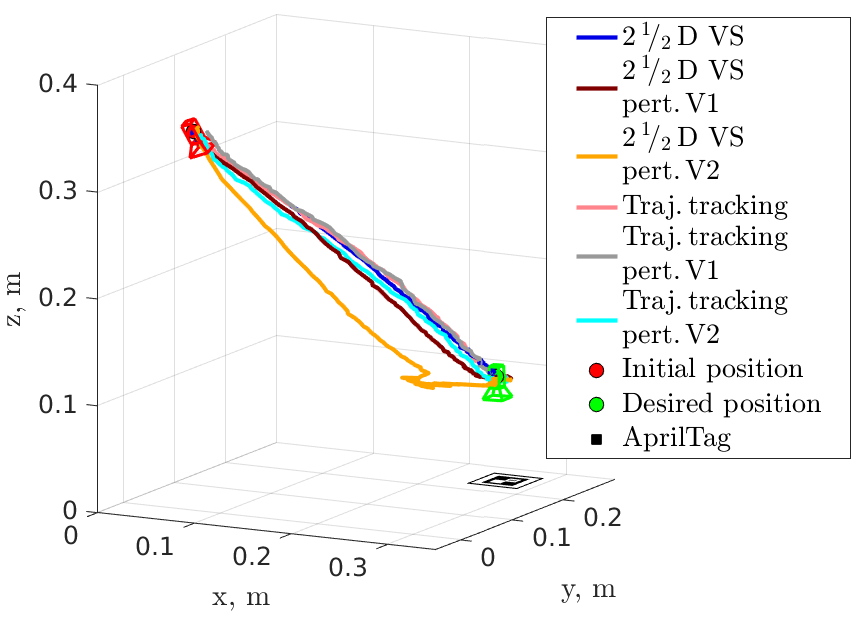}}%
\hfil
\subfloat[\label{subd}]{\includegraphics[height = 6cm]{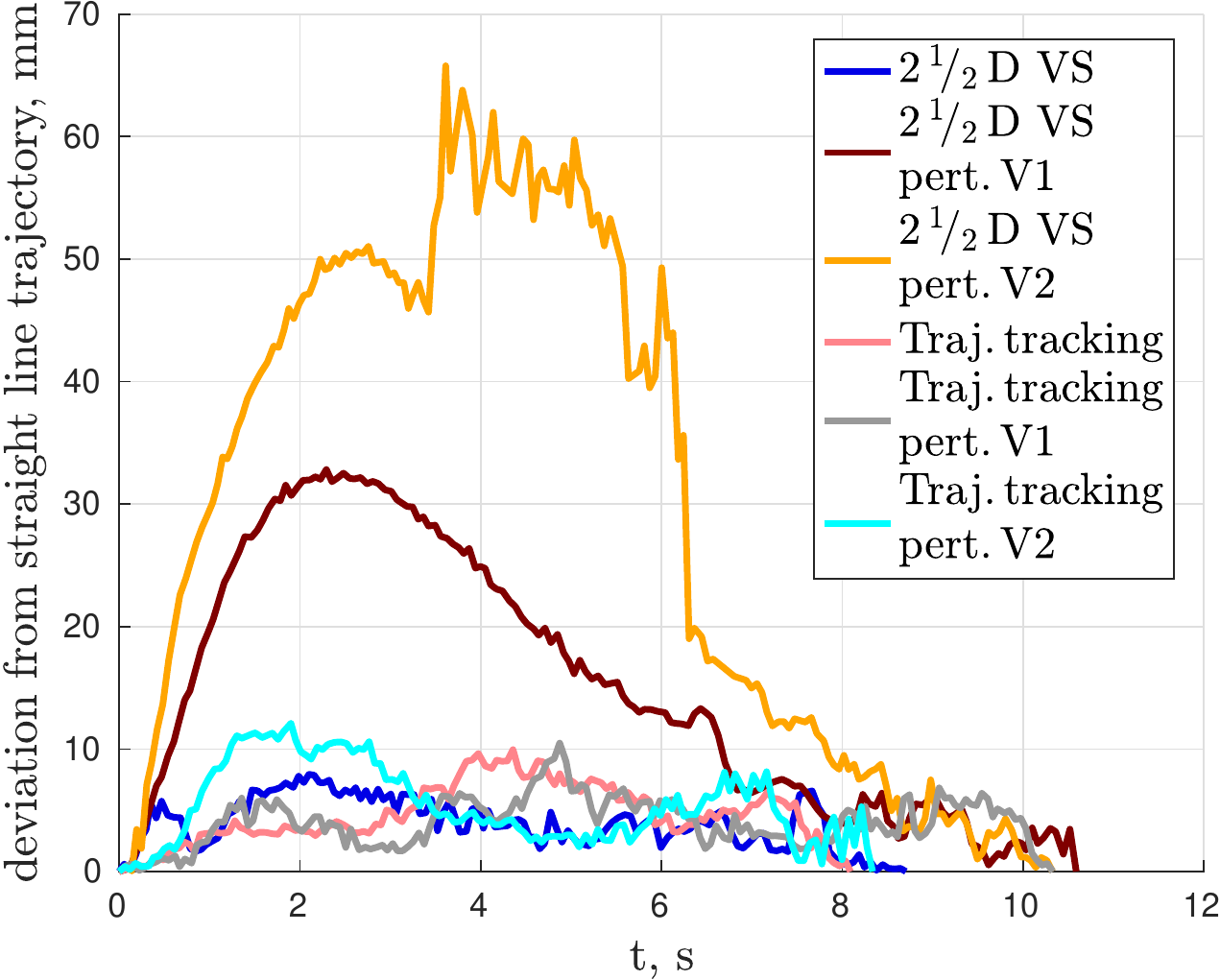}}%
 \caption{\twohalfdvs\ experiments on ACROBOT: a) the trajectory of AprilTag center-point in the image; b) The pixel deviation from the ideal straight-line trajectory; c) the trajectory of the camera in the frame~$\F{b}$; d) the deviation from the ideal straight-line 3D trajectory. }
\label{fig:figuresss}
\end{figure*}

\begin{figure}[thpb]
      \centering
      \includegraphics[width = 1\columnwidth]{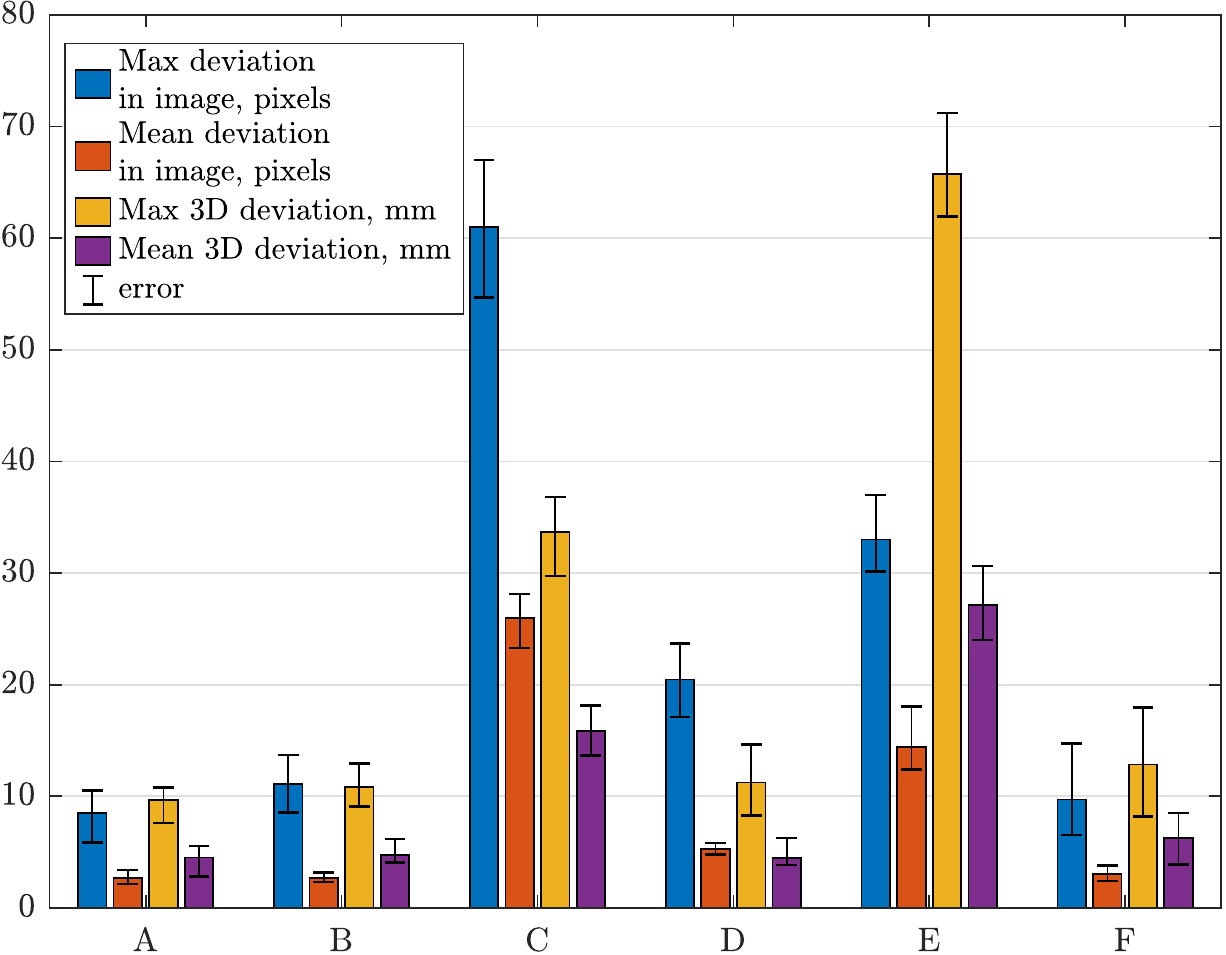}
      \caption{Bar graph showing the max and mean deviation from the ideal object center-point $\mbf{c}$ trajectory in image and the ideal camera trajectory in $\F{b}$ with and without voluntarily added perturbations. Classical \twohalfdvs\ without added perturbation (A), under the effect of perturbation set $V1$ (C) and $V2$ (E);
       \twohalfdvs\ with trajectory tracker without added perturbation (B), under the effect of perturbation set $V1$ (D) and $V2$ (F).
       }
      \label{fig:bargraph}
   \end{figure}

Figure~\ref{fig:figuresss} shows the experimental results (see also the accompanying video). Figure~\ref{suba} shows the trajectories of the AprilTag center-point in the image, while Fig.~\ref{subc} shows the 3D trajectories of the camera in the frame~$\F{b}$. Additionally, the deviation from the straight-line trajectory in the image and in~$\F{b}$ is shown in Figs.~\ref{subb} and~\ref{subd}, respectively. Each controller, the classic \twohalfdvs\ and the one with trajectory tracking (named ``Traj.\,tracking" in Fig.~\ref{fig:figuresss}) was tested without added perturbations and under the effect of each perturbation set~$V1$ and~$V2$. Each experiment was repeated 15 times and the results are combined in a bar graph shown in Fig.~\ref{fig:bargraph}.

Under good conditions, the behavior is as expected, namely, we see straight-line trajectories both in~3D and in the image.
When no perturbation is added, the behavior of \twohalfdvs\ controller with and without trajectory tracking is similar. For both controllers the deviation does not surpass~0.01\,m and 10 pixels. 
 The superiority of trajectory tracking can be clearly seen when the system is perturbed. Each of the perturbation sets forces the classic \twohalfdvs\ to produce deviations from the ideal trajectories. $V2$ leads to higher deviation on the 3D trajectory (orange line in Fig.~\ref{subd}), while $V1$ has a more pronounced effect on the trajectory in the image (brown line in Fig.~\ref{subb}). 
On the contrary, the perturbation sets have a minimal effect on the trajectories produced by the controller with trajectory tracker as depicted by the gray and cyan lines in Fig.~\ref{fig:figuresss} for $V1$ and $V2$, resp. Indeed, for the 3D trajectory three lines corresponding to the trajectory tracking controller remain very near. The behavior is slightly worse in the image, where perturbation set $V1$ leads to about 18 pixel error (gray line). However, it is three times smaller than the almost 55 pixel error (brown line) obtained with the classic \twohalfdvs\ under the same perturbations in Fig.~\ref{subb}.

Figure~\ref{fig:bargraph} shows the max and mean deviation from the ideal 2D and 3D trajectories for both controllers subject to the three perturbation sets. When there is no perturbation, the behavior of the controller without and with trajectory tracker is similar (groups~A and~B). No matter the perturbation set, the errors are at least three times smaller when the trajectory tracker is used: groups~C and~D for~$V1$; groups~E and~F for~$V2$. Furthermore, the 3D trajectory deviation (and the deviation of the trajectory in image for $V2$) remains similar to the trajectory tracker without perturbation.

\section{Conclusions}
\label{s6}

This paper dealt with the use of trajectory planning and tracking with 2\textonehalf D~Visual Servoing for the control of Cable-Driven Parallel Robots. First, the proposed controller aims to increase the robustness of the system with respect to perturbations and errors in the robot model. Furthermore, it ensures the straight-line motion of  both the center-point of the AprilTag in the image and the camera in the base frame.

Furthermore, a Control Stability Workspace~(CSW) was defined and computed for a CDPR prototype ACROBOT, based on the stability analysis of the full system under {2\textonehalf D}~visual servoing control.
The effect of perturbations on CSW size was highlighted.

The improvement of robustness due to the use of trajectory planning and tracking was clearly shown in experimental validation. While both systems, namely, without and with trajectory tracking, remain stable and achieve the set goal, the trajectory produced by the former is clearly affected by perturbations.

A further improvement would be developing a control law that allows us to detect and counteract the modeling errors, instead of increasing robustness to these errors.


\end{document}